\documentclass[letterpaper, 10 pt, conference]{ieeeconf}  % Comment this line out if you need a4paper

\IEEEoverridecommandlockouts                              % This command is only needed if 
\usepackage{graphicx}
\usepackage{color,soul}
\usepackage{dblfloatfix}
\usepackage{multirow}
\usepackage{multicol}
\usepackage{hyperref}

\begin{document}
\title{\LARGE \bf 
Augmented Reality Remote Operation of Dual\\Arm
Manipulators in Hot Boxes}

\author{\protect\parbox{\textwidth}{\protect\centering Frank Regal$^{1}$, Young Soo Park$^{2}$, Jerry Nolan$^{2}$, and Mitch Pryor$^{1}$\\}
% <-this % stops a space
\thanks{$^{1}$The University of Texas at Austin,
        Austin, TX 78712, USA
        {\tt\small fregal@utexas.edu}}% <-this % stops a space
\thanks{$^{2}$Argonne National Laboratory,
        Lemont, IL 60439, USA
        {\tt\small ypark@anl.gov}}%
}
\maketitle

\section{INTRODUCTION}
In nuclear isotope and chemistry laboratories, hot cells and gloveboxes provide scientists with a controlled and safe environment to perform experiments \cite{osti_4327593}. Working on experiments in these isolated containment cells requires scientists to be physically present. For hot cell work, scientists manipulate equipment and radioactive material inside through a bilateral mechanical control mechanism. Motions produced outside the cell with the master control levers are mechanically transferred to the internal grippers inside the shielded containment cell. Similar to hot cells, gloveboxes provide scientists with the capability to manipulate equipment inside, but instead, the users can perform work with their hands by using sealed glove inserts on the side of the chamber.

There is a growing need to have the capability to conduct experiments within these cells remotely. During past pandemics, scientists often could not enter facilities and had to restart their time-sensitive work. In other situations, scientists' home sites often do not have all equipment needed for the experiment that other sites might have. Therefore, scientists need a solution to perform physical experiments outside the laboratory.

\begin{figure}[b]
    \centering
    \includegraphics[width=\linewidth]{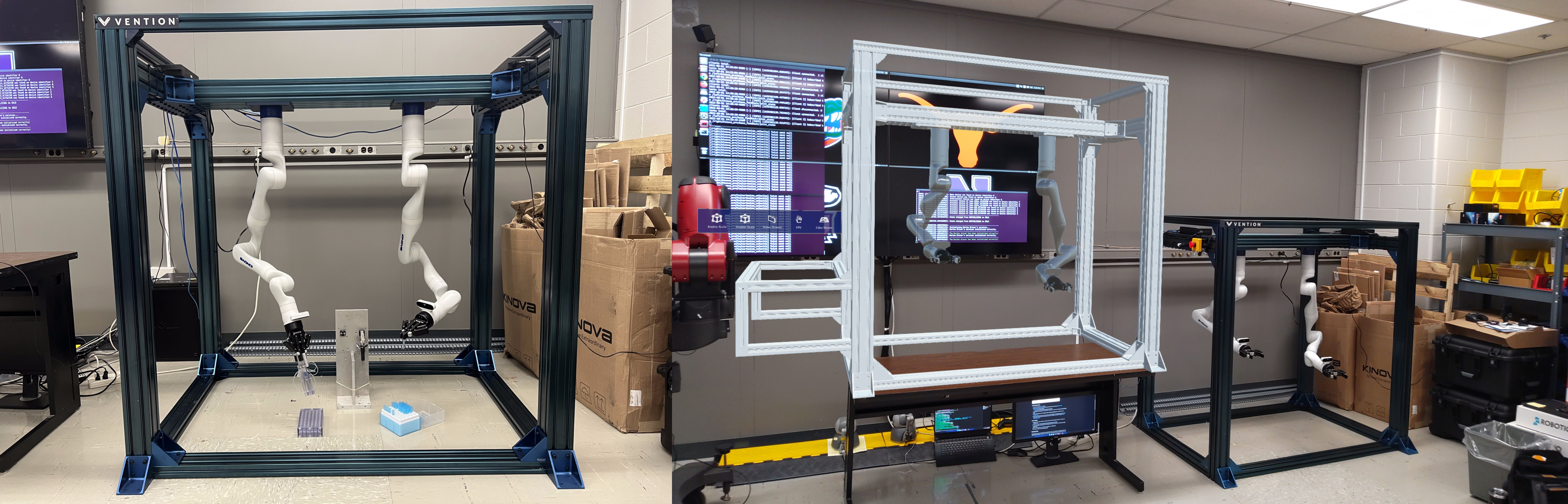}
    \caption{\textbf{[Left] New hot box design with two robotic arms mounted to the top manipulating chemistry tools. [Right] First-person-view of a HoloLens 2 user seeing both a virtual replica of the hot box atop the table and the real physical hotbox on the right.}}
    \label{fig:real_physical_hot_box}
\end{figure}

Remote robotic control has long been a significant area of research. Robotic teleoperated systems have assisted in natural disaster clean-up, hazardous environment inspections, and nuclear facility decommissioning with success \cite{siciliano_springer_2016}, \cite{sudevan_current_2018}.
A simple method to enable remote manipulations within hot cell and glovebox cells is to mount two robotic arms inside a box to mimic the motions of human hands. An initial prototype, coined the \textit{hot box}, has been built and is shown in Figure \ref{fig:real_physical_hot_box}. Capable manipulators can perform structured pre-defined experiments autonomously without issue with the robotic arms programmed to fit the experiment. Problems arise when the experimental workspace within the cell changes. Enabling a robot to understand, plan, and act in an unstructured ever-changing experimental environment in a fully autonomous manner is difficult to repeat and not robust enough to be deployed for laboratory applications where the task is not known before hand or may only be undertaken a single time. Therefore, teleoperation of the robotic arms is required even if automation is possible. Teleoperation will always be a necessary capability for safety purposes. Scientists need to operate the dual arm, robotic manipulators, in the \textit{hot box} with intuitive and natural controls and be fully situationally aware of their experiments.

Current solutions to perform remote teleoperation of robotic systems use 2-dimensional computer monitors with multiple windows, viewports, buttons, joysticks, and keyboards that often overload untrained operators \cite{marturiAdvancedRoboticManipulation2016}, \cite{sharp_semiautonomous_2017}.

Augmented Reality (AR) headsets, such as the Microsoft HoloLens 2, have been used to connect human operators with robotic agents to communicate and command mobile robotic platforms \cite{regalAugre2022}, \cite{delmerico_spatial_2022}. \cite{sita_towards_2017} paired a HoloLens with an industrial manipulator to allow users to visualize planned joint states and provide the arm with goal poses via manipulation of a virtual cube. Manring et al. \cite{dervilis_special_2020} used a HoloLens to allow users to manipulate a virtual representation of a robotic arm, visualize the planned motion in the AR environment, and then send execution commands to the manipulator to proceed.

Building upon work from \cite{sita_towards_2017} and \cite{dervilis_special_2020}, our team at Argonne National Laboratory developed an AR application for a HoloLens 2 headset that enables users to teleoperate \textit{hot box} dual arm manipulators by grasping robotic end-effectors of digital replicas in AR from a remote location. In addition to the real-time replica of the physical robotic arms in AR, the application also enabled users to view a live video stream attached to the robotic arms and a parsed 3D point cloud of 3D objects in their remote AR environment for better situational awareness. It also provides users with virtual fixtures to assist in manipulation and other teleoperation tasks.

\begin{figure}[b!]
    \centering
    \includegraphics[width=\linewidth]{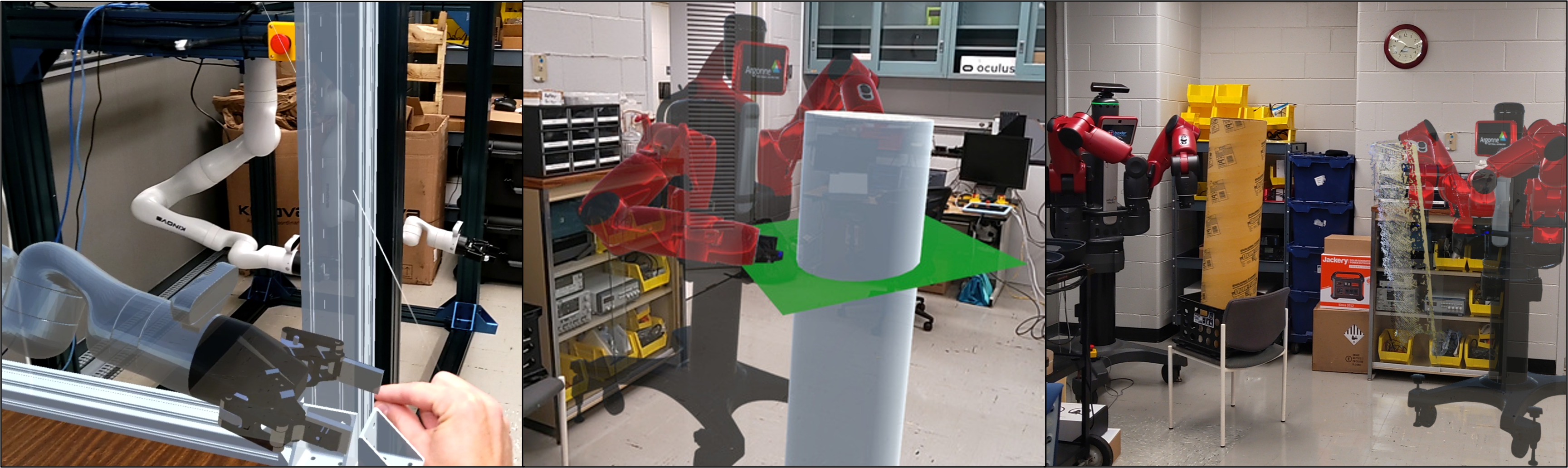}
    \caption{\textbf{Images captured from the perspective of a HoloLens 2 user. [Left] User pinching the virtual end effector commanding the physical arm to move. [Middle] Robotic arm interacting with a virtual plane in AR used to assist in teleoperation. [Right] Virtual point cloud visualization of the physical tube.}}
    \label{fig:appkication_capabilities}
\end{figure}

\begin{figure}[ht!]
    \centering
    \includegraphics[width=\linewidth]{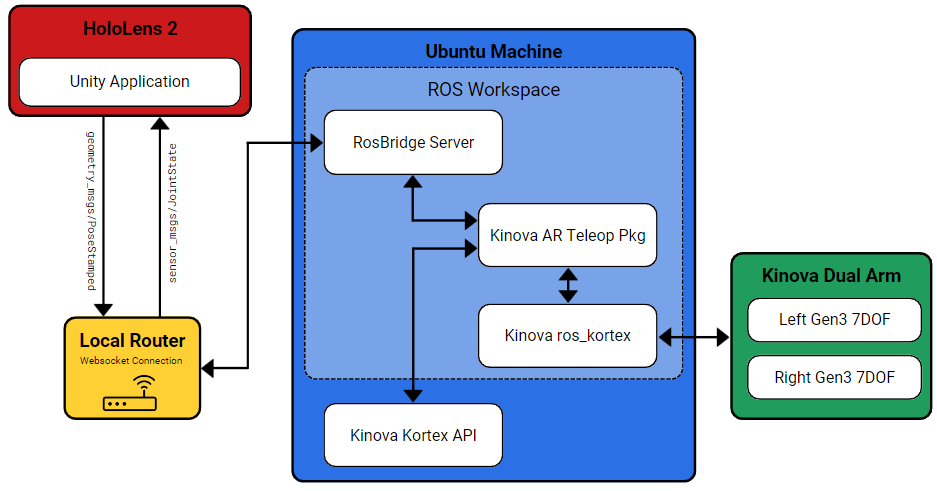}
    \caption{\textbf{System overview of how the HoloLens 2 integrates with the robotic manipulators inside of the hot box.}}
    \label{fig:system_overview}
\end{figure}

\section{SYSTEM OVERVIEW} \label{sec:sys_arc}
A Microsoft HoloLens 2, ROS packages, and Kinova Robotic's Gen3 robotic arms were combined to create a framework that allows remote operation via natural controls. Unity game engine and ROS package scripts created the communication pipeline required to communicate back and forth between the headset and robotic arms. With the augmented reality application scene setup, a user can manipulate the robotic arms with a pinch, using their thumb and index finger on the virtual arm grippers. The HoloLens 2 tracks the pose of the hand positioning. Those hand poses convert to serialized messages and are sent to an Ubuntu machine over a wireless network acting as a control station server. This server deserializes the messages and broadcasts them to specific scripts running on the server and robotic arms that interpret the messages and converts them to twist commands. When the robots move their position, the HoloLens 2 receives those location messages of the robotic arms through the reverse route, and the virtual arms within the HoloLens 2 send updates accordingly. Through this configuration, a user has an experience that allows them to see precisely the position of the robotic arms in the \textit{hot box} from remote locations and the ability to control them effortlessly.

\subsection{HoloLens 2} 
As shown in Figure 3, the Unity game engine was used to build the AR application. The game engine application provides HoloLens
2 users with virtual representations of the \textit{hot box} and read the hand tracking data from the HoloLens. The Unity application on the HoloLens 2 publishes the hand positions and subscribes to the robotic arms' joint positions on ROS topics published on a RosBridge server. These ROS topic messages are standard \textit{PoseStamped} and \textit{JointState} ROS messages for the hands and robot joint positions, respectively. Before the Unity application sends the messages over the wireless network to the server, the Unity application serializes the messages into a standard JSON format. This JSON format allows for rapid translation of data across a local router.

\subsection{Linux Machine}
When the serialized ROS messages leave the HoloLens 2, a ROS package called RosBridge, running on a Linux server, begins to receive these messages, deserializes them, and broadcasts them locally to the ROS nodes on the Linux machine. One specific node is the Kinova AR Teleop Package. This custom ROS package takes the \textit{PoseStamped} positions of the hands that the HoloLens published and converts them to a linear and angular velocity that the robots understand, called a \textit{Twist} message. After converting to a \textit{Twist} message, the Kinova AR Teleop Package publishes the new \textit{Twist} message to the Kinova ROS Kortex package. This package is the low-level ROS package that sends the move commands to the robots. The ROS Kortex takes the \textit{Twist} messages, calls the internal inverse kinematic solver, and sends a move command to the arms. Once the arms move, the ROS Kortex package takes the continuously published \textit{JointState} information and publishes this back to the Kinova AR Teleop Package. The messages are then packaged, serialized, and sent back to the HoloLens 2 device across the network. The HoloLens 2 finally interprets the message, deserializes, and then updates the virtual representation position of the holographic Kinova arms in the Hololens 2 so that the user can see its new position from a remote location.

\section{TEST IMPLEMENTATION} \label{sec:applications}
The application has been used to control both the Kinova dual arm and a ReThink Robotics Baxter dual arm. Users can move the manipulators with a pinch and grab, from a remote location. A live video stream, virtual fixtures, and point cloud visualizations can be toggled within the interface. The virtual robots can also scale in their scene for precise manipulations or confined space arrangements.

\section{CONCLUSION} \label{sec:results}

Within this work, an AR application was created to allow scientists to remotely operate robotic manipulators via hand gestures inside a \textit{hot box} to enable scientists to work remotely in a hot cell or glovebox environment. A live video stream, 3D point cloud visualization, and virtual fixtures have been tested within the AR environment as well. Future research needs to be done to understand the best hand gesturing method to control the robotic arms’ position and orientation. Currently, users are using pinch and drag, but with the HoloLens 2, much more sensor data about the human user is captured. There could be more research done to understand if inputs like voice, eye gaze, and head position could be used to improve the users’ remote operation experience.

\vspace{7pt}
\break
\centering \textit{Demonstration Video}\\
$\centering \href{https://youtu.be/aPauLc2_o00}{https://youtu.be/aPauLc2_o00}$

%%%%%%%%%%%%%%%%%%%%%%%%%%%%%%%%%%%%%%%%%%%%%%%%%%%%%%%%%%%%%%%%%%%%%%%%%%%%%%%%

% \begin{thebibliography}{99}
\typeout{}
\bibliographystyle{IEEEtran}
\bibliography{IEEEabrv,bibliography}

\begin{thebibliography}{1}
\providecommand{\url}[1]{#1}
\csname url@rmstyle\endcsname
\providecommand{\newblock}{\relax}
\providecommand{\bibinfo}[2]{#2}
\providecommand\BIBentrySTDinterwordspacing{\spaceskip=0pt\relax}
\providecommand\BIBentryALTinterwordstretchfactor{4}
\providecommand\BIBentryALTinterwordspacing{\spaceskip=\fontdimen2\font plus
\BIBentryALTinterwordstretchfactor\fontdimen3\font minus
  \fontdimen4\font\relax}
\providecommand\BIBforeignlanguage[2]{{%
\expandafter\ifx\csname l@#1\endcsname\relax
\typeout{** WARNING: IEEEtran.bst: No hyphenation pattern has been}%
\typeout{** loaded for the language `#1'. Using the pattern for}%
\typeout{** the default language instead.}%
\else
\language=\csname l@#1\endcsname
\fi
#2}}

\bibitem{osti_4327593}
\BIBentryALTinterwordspacing
E.~R. Fosdick, ``Aec hot cells and related facilities,'' 5 1958. [Online].
  Available: \url{https://www.osti.gov/biblio/4327593}
\BIBentrySTDinterwordspacing

\bibitem{siciliano_springer_2016}
\BIBentryALTinterwordspacing
``Springer handbook of robotics.'' [Online]. Available:
  \url{https://link.springer.com/10.1007/978-3-319-32552-1}
\BIBentrySTDinterwordspacing

\bibitem{sudevan_current_2018}
V.~Sudevan, A.~Shukla, and H.~Karki, ``Current and future research focus on
  inspection of vertical structures in oil and gas industry,'' in \emph{2018
  18th International Conference on Control, Automation and Systems ({ICCAS})},
  pp. 144--149.

\bibitem{marturiAdvancedRoboticManipulation2016}
N.~Marturi, A.~Rastegarpanah, C.~Takahashi, M.~Adjigble, R.~Stolkin, S.~Zurek,
  M.~Kopicki, M.~Talha, J.~A. Kuo, and Y.~Bekiroglu, ``Towards advanced robotic
  manipulation for nuclear decommissioning: {{A}} pilot study on tele-operation
  and autonomy,'' in \emph{2016 {{International Conference}} on {{Robotics}}
  and {{Automation}} for {{Humanitarian Applications}} ({{RAHA}})}, Dec. 2016,
  pp. 1--8.

\bibitem{sharp_semiautonomous_2017}
A.~Sharp, K.~Kruusamäe, B.~Ebersole, and M.~Pryor, ``Semiautonomous dual-arm
  mobile manipulator system with intuitive supervisory user interfaces,'' in
  \emph{2017 {IEEE} Workshop on Advanced Robotics and its Social Impacts
  ({ARSO})}, pp. 1--6.

\bibitem{regalAugre2022}
F.~Regal, C.~Petlowany, C.~Pehlivanturk, C.~V. Sice, C.~Suarez, B.~Anderson,
  and M.~Pryor, ``Augre: Augmented robot environment to facilitate human-robot
  teaming and communication,'' in \emph{2022 31st {{IEEE Int. Symposium}} on
  {{Robot}} and {{Human Interactive Com.}} ({{RO-MAN}})}, August 2022.

\bibitem{delmerico_spatial_2022}
J.~Delmerico, R.~Poranne, F.~Bogo, H.~Oleynikova, E.~Vollenweider, S.~Coros,
  J.~Nieto, and M.~Pollefeys, ``Spatial computing and intuitive interaction:
  Bringing mixed reality and robotics together,'' vol.~29, no.~1, pp. 45--57,
  conference Name: {IEEE} Robotics \& Automation Magazine.

\bibitem{sita_towards_2017}
\BIBentryALTinterwordspacing
E.~Sita, M.~Studley, F.~Dailami, A.~Pipe, and T.~Thomessen, ``Towards
  multimodal interactions: robot jogging in mixed reality,'' in
  \emph{Proceedings of the 23rd {ACM} Symposium on Virtual Reality Software and
  Technology}, ser. {VRST} '17.\hskip 1em plus 0.5em minus 0.4em\relax
  Association for Computing Machinery, pp. 1--2. [Online]. Available:
  \url{https://doi.org/10.1145/3139131.3141200}
\BIBentrySTDinterwordspacing

\bibitem{dervilis_special_2020}
\BIBentryALTinterwordspacing
``Special topics in structural dynamics \& experimental techniques, volume 5:
  Proceedings of the 37th {IMAC}, a conference and exposition on structural
  dynamics 2019.'' [Online]. Available:
  \url{http://link.springer.com/10.1007/978-3-030-12243-0}
\BIBentrySTDinterwordspacing

\end{thebibliography}

% \bibitem{c1} G. O. Young, ÒSynthetic structure of industrial plastics (Book style with paper title and editor),Ó 	in Plastics, 2nd ed. vol. 3, J. Peters, Ed.  New York: McGraw-Hill, 1964, pp. 15Ð64.

% \end{thebibliography}

\end{document}